%% file: 0-main.tex
\RequirePackage{fix-cm}
\documentclass[twocolumn]{svjour3}          % twocolumn
\smartqed  % flush right qed marks, e.g. at end of proof
\usepackage{booktabs}
\usepackage{amsmath}
\usepackage{amssymb}
\usepackage{mathtools}
\usepackage{xspace}
\usepackage{hyperref}
\usepackage{graphicx}
\usepackage{caption}
\usepackage{subcaption}
\usepackage{multirow}
\usepackage{color, colortbl}
\usepackage{fancyvrb}
\usepackage{xcolor}
\usepackage{appendix}
\usepackage{natbib}

\newcommand{\eg}{\emph{e.g.,}\xspace}

\newcommand{\ie}{\emph{i.e.,}\xspace}

\DeclareMathOperator*{\argmin}{arg\,min}

\newcommand{\reb}[1]{\textcolor{black}{#1}}
\newcommand{\nreb}[1]{\textcolor{black}{#1}}
\newcommand{\nnreb}[1]{\textcolor{black}{#1}}

\definecolor{Gray}{gray}{0.9}
\definecolor{pink}{rgb}{1.0, 0.85, 0.85}

\journalname{International Journal of Computer Vision}

\hypersetup{pdfstartview={FitH}}
\hypersetup{pdfborder={0 0 0}}
\hypersetup{pdfpagemode={UseNone}}
\hypersetup{colorlinks=true}
\hypersetup{citecolor=blue}
\hypersetup{linkcolor=blue}

\begin{document}

\title{Universal Prototype Transport for Zero-Shot Action\\Recognition and Localization}

\author{Pascal Mettes
}

\institute{Pascal Mettes \at
Universiteit van Amsterdam, Amsterdam, the Netherlands\\
\email{P.S.M.Mettes@uva.nl}}

\date{Received: date / Accepted: date}

\maketitle

\input{1-abstract}
\input{2-introduction}
\input{3-relatedwork}
\input{4-method}
\input{5-experiments}
\input{6-conclusions}

\bibliographystyle{spbasic}      % basic style, author-year citations
\bibliography{egbib-new}   % name your BibTeX data base

\end{document}

%% file: 1-abstract.tex
\begin{abstract}
This work addresses the problem of recognizing action categories in videos when no training examples are available. The current state-of-the-art enables such a zero-shot recognition by learning universal mappings from videos to a semantic space, either trained on large-scale seen actions or on objects.
%While effective, we find that universal action and object mappings are biased to their seen categories. Such biases are further amplified due to biases between seen and unseen categories in the semantic space.
While effective, we find that universal action and object mappings are biased to specific regions in the semantic space.
These biases lead to a fundamental problem: many unseen action categories are simply never inferred during testing. For example on UCF-101, a quarter of the unseen actions are out of reach with a state-of-the-art universal action model. To that end, this paper introduces universal prototype transport for zero-shot action recognition. The main idea is to re-position the semantic prototypes of unseen actions by matching them to the distribution of all test videos. For universal action models, we propose to match distributions through a hyperspherical optimal transport from unseen action prototypes to the set of all projected test videos. The resulting transport couplings in turn determine the target prototype for each unseen action. Rather than directly using the target prototype as final result, we re-position unseen action prototypes along the geodesic spanned by the original and target prototypes as a form of semantic regularization.
%We then define a target prototype for each unseen action as the weighted Fr\'echet mean over the transport couplings. Equipped with a target prototype, we re-position unseen action prototypes along the geodesic spanned by the original and target prototypes, acting as a form of semantic regularization.
For universal object models, we outline a variant that defines target prototypes based on an optimal transport between unseen action prototypes and object prototypes. 
\nnreb{Empirically, we show that universal prototype transport diminishes the biased selection of unseen action prototypes and boosts both universal action and object models for zero-shot classification and spatio-temporal localization.}
%Empirically, we show that universal prototype transport diminishes the biased selection of unseen action prototypes and boosts both universal action and object models, resulting in state-of-the-art performance for zero-shot classification and spatio-temporal localization.
\end{abstract}

%% file: 2-introduction.tex
\section{Introduction}
This paper addresses the problem of recognizing actions in videos. Foundational deep network approaches performed action recognition through frame-level fusion \citep{karpathy2014large}, two-stream networks \citep{simonyan2014two,feichtenhofer2016convolutional}, and 3D convolutional networks~\citep{carreira2017quo}. Building upon these approaches, recent works have shown great recognition capabilities through \eg slow-fast architectures~\citep{feichtenhofer2019slowfast}, separated 3D convolutions~\citep{tran2019video}, and video transformers~\citep{arnab2021vivit}. Such deep networks require large amounts of video material for training and efforts have been made to meet those video demands, such as ActivityNet \citep{caba2015activitynet}, EPIC Kitchens \citep{damen2018scaling}, Kinetics \citep{carreira2017quo}, HowTo100M \citep{miech2019howto100m}, and EGO4D \citep{grauman2022ego4d} to name a few. While such datasets increase the coverage of the action space, we seek to recognize actions even when no examples are available during training.

In zero-shot action recognition, many works have outlined approaches that mirror successes in the image domain, for example by using attributes~\citep{liu2011recognizing,gan2016learning} or feature synthesis~\citep{mishra2020zero} to transfer knowledge from seen to unseen actions. More recently, state-of-the-art results have been achieved by taking a universal learning perspective, where large-scale models are trained to map input videos to a shared semantic space occupied by both seen and unseen categories. In the first universal perspective, large-scale networks train a mapping from videos to a semantic space on hundreds of seen actions from \eg ActivityNet~\citep{zhu2018towards} or Kinetics~\citep{brattoli2020rethinking,pu2022alignment}. For a target dataset with unseen actions, zero-shot inference is directly possible through a nearest neighbour search between the video mappings and the embeddings of unseen actions. In the second perspective, networks are trained on thousands of objects~\citep{jain2015objects2action,mettes2021object}, where inference becomes a function of the object likelihoods in test videos and the semantic relation between objects and actions.

\begin{figure}[t]
\centering
\includegraphics[width=\linewidth]{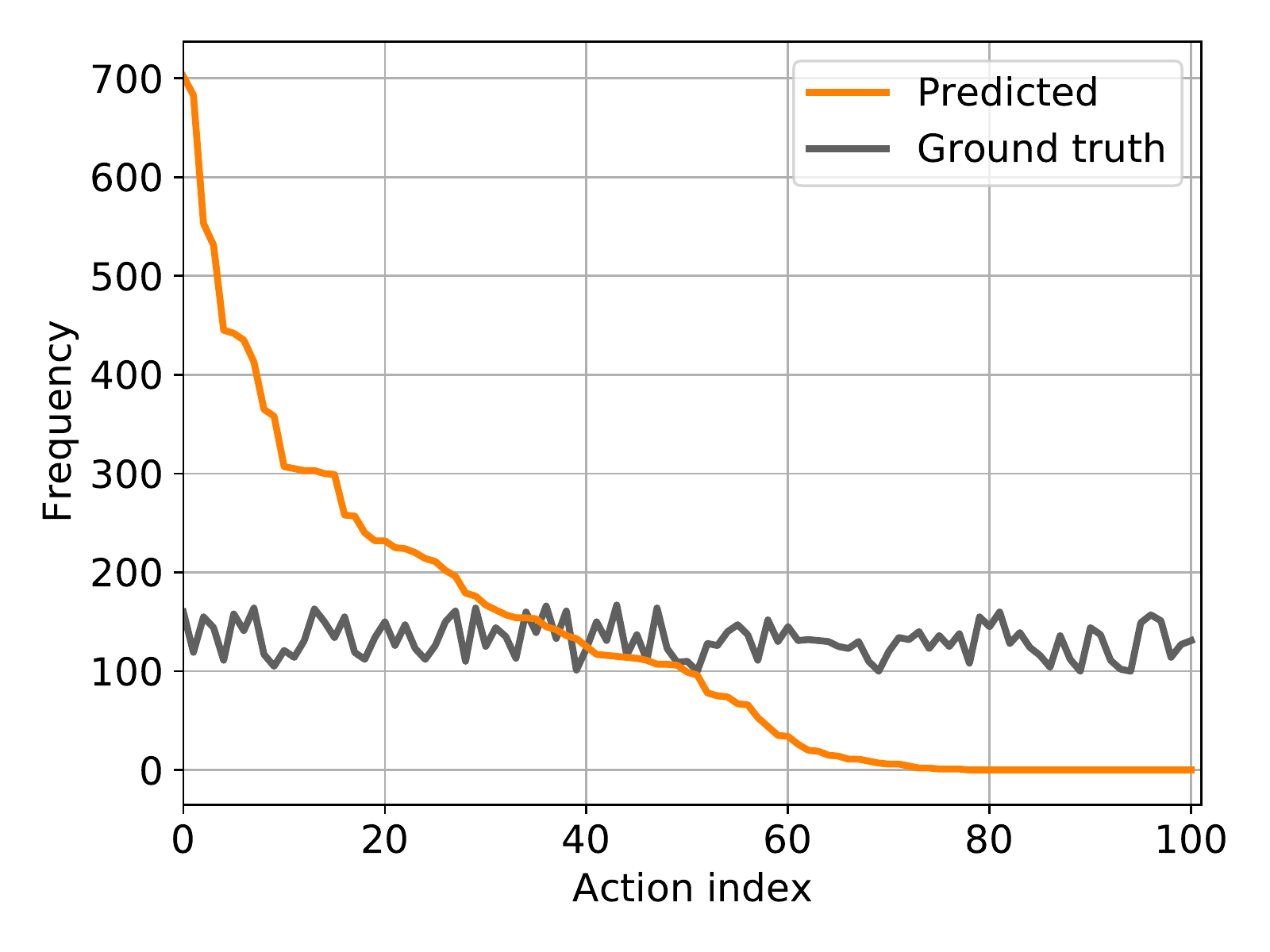}
\caption{\textbf{Predicted and ground truth} action distributions for an example universal action model. The distributions are out of alignment, with a quarter of the actions never predicted altogether, severely limiting the upper bound for zero-shot action recognition. We seek to overcome this misalignment by matching the distributions of unseen actions and test videos.}
\label{fig:fig1}
\end{figure}

%While effective, both universal perspectives share a common limitation: they are strongly biased to subsets of the unseen actions. Compounding biases in (i) the mapping of videos to seen categories and (ii) in the matching between seen and unseen categories result in a mismatch between the prototypes of unseen actions and projected test videos. A significant part of the actions are as a result simply never selected, disrupting progress in zero-shot action recognition. Figure~\ref{fig:fig1} illustrates this problem on UCF-101 for example with the recent universal action model of \cite{brattoli2020rethinking}. Here, 23\% (!) of the unseen actions are simply never selected for any test video in the first place, placing hard upper bounds on the accuracy in zero-shot recognition. We seek to address this limitation.
\reb{While effective, both universal perspectives share a common limitation: they are strongly biased to subsets of the unseen actions. Compounding biases in (i) the mapping of videos to seen categories and (ii) in the matching between seen and unseen categories result in a mismatch between the prototypes of unseen actions and projected test videos. A significant part of the actions are as a result simply never selected, disrupting progress in zero-shot action recognition. This is a direct result of their inductive nature, where each test video is individually evaluated. Figure~\ref{fig:fig1} illustrates this problem on UCF-101 for example with the recent universal action model of \cite{brattoli2020rethinking}. Here, 23\% (!) of the unseen actions are simply never selected for any test video in the first place, placing hard upper bounds on the accuracy in zero-shot recognition. We seek to address this limitation by enriching universal models with an optimal transport transductive perspective.}

We introduce universal prototype transport for zero-shot action recognition with universal models. The main idea is to re-position the prototypes of unseen actions in the shared semantic space to better align with the test videos. For universal action models, we first find an optimal mapping from action prototypes to test videos on the hypersphere. We then define a target prototype for each unseen action as a weighted Fr\'{e}chet mean based on the coupling matrix from the optimal transport. Rather than doing a nearest neighbor search directly on the target prototypes for zero-shot recognition, we re-position the prototypes through interpolation along the geodesic spanned by the original and target prototypes. The intuition behind the interpolation is to maintain a form of semantic regularization for the target prototypes from their original semantic representation. For universal object models, we follow a similar setup, but replace the test videos with object prototypes during the optimal transport mapping. Beyond action recognition, we show that our approach is also helpful for zero-shot spatio-temporal action localization due to an improved ranking of tubes from different test videos to unseen actions.

We perform empirical evaluations on four action datasets for two tasks: zero-shot action recognition and zero-shot spatio-temporal action localization. The experiments confirm that universal prototype transport diminishes the biased selection of unseen actions in universal models, resulting in better recognition performance.
\reb{By combining inductive universal action and object models, we are able to improve both zero-shot action recognition and spatio-temporal action localization in videos.}
%By building upon and combining universal action and object models, we obtain new state-of-the-art results for recognition and spatio-temporal localization.
Our approach is general in nature and can be used on top of any existing approach. The code will be made available on github.

%% file: 3-relatedwork.tex
\section{Related work}

\subsection{Zero-shot action recognition}

%\psmm{To do: add references from 2022}

Zero-shot action recognition refers to the task of assigning an action label to a test video given a pool of actions not observed during training. A common approach is to learn and transfer a shared representation on seen actions with training examples to be able to perform inference on unseen actions. A well-known shared space is given by attributes~\citep{gan2016learning,liu2011recognizing,zhang2015robust}. By projecting test videos to an attribute space, inference is possible through a neighbour search with unseen actions manually defined in the same space. Since attributes require manual annotations for every action, other works prefer to use word embeddings~\citep{bishay2019tarn,gan2016concepts,gan2016recognizing,li2016recognizing}, video captions~\citep{estevam2021tell}, or action hierarchies~\citep{long2020searching} to provide a shared space for seen and unseen actions.

State-of-the-art zero-shot action recognition solutions take a universal learning perspective with semantic word embeddings as the shared space for knowledge transfer. Rather than relying on a small set of seen actions from the same dataset, large-scale models are trained on hundreds or thousands of seen categories to learn a direct mapping from videos to the shared space. The first universal learning direction relies on large-scale actions with training videos. \cite{zhu2018towards} were the first to propose a large-scale universal action perspective by learning a video network on 200 actions from ActivityNet~\citep{caba2015activitynet}. \cite{brattoli2020rethinking} have obtained high performance in zero-shot action recognition by scaling this perspective to 664 actions from Kinetics~\citep{carreira2017quo}. \cite{pu2022alignment} have subsequently shown that incorporating alignment, uniformity, and feature synthesis further improve recognition. Due to the large amount of seen actions, care needs to be taken to avoid (near) duplicates between seen and unseen actions, as also noted by \cite{roitberg2018towards}. These universal models share a similar fate, where the predicted and ground truth action distributions are mis-aligned due to biased training. We show that our proposed prototype transport improves the current state-of-the-art action-based models for zero-shot recognition.

Next to universal action models, several works have shown competitive results by taking a universal object perspective. Large-scale networks are trained in the image domain on thousands of object labels~\citep{jain2015objects2action,liu2019generalized,mettes2017spatial,mettes2021object,wu2016harnessing}. Once trained any action can be inferred based on the object likelihoods in test video and the semantic similarities between objects and unseen actions. \nreb{For example, \citet{bretti2021zero} show that actions like skateboarding are easy to recognize as its corresponding action embedding is close to object embeddings like skateboard and roller skates.} Similar to universal action models, we show that the object-based perspective also benefits from our proposed approach.
%Regardless of the universal perspective, the current state-of-the-art shares the same fate: zero-shot inference results in a long-tailed distribution over selected actions. Many unseen actions are simply never selected due to persistent biases. We tackle this problem with a prototype transport over universally trained models. This paper introduces a new way of dealing with biases in universal models, hence the focus on the video domain over the image domain, where universal models are not state-of-the-art.

%Within zero-shot action recognition, a transductive view is commonly applied, as noted by~\cite{estevam2021zero}.
\nnreb{By operating on the entire test distribution, we take a transductive view, a common setting in zero-shot action recognition as noted by \cite{estevam2021zero}.
Where inductive reasoning requires solving a general problem and applying them to individual samples, transduction is about reasoning from observed training cases to observed test cases. It is hence commonly seen as a more direct way to solve inference problems \citep{vapnik2006estimation} with direct implications for zero-shot learning, semi-supervised learning, transfer learning, and more \citep{arnold2007comparative}. In practical settings, inference is often performed on video collections, for example for recommendation or searching in large databases, making transductive learning a viable learning setting.}
\cite{rohrbach2013transfer} provide a foundation for transduction in zero-shot context by exploiting the inter-sample similarity over the test set. In similar spirit, several works have proposed transductive extensions for zero-shot action recognition~\citep{alexiou2016exploring,fu2014transductive,gao2019know,kodirov2015unsupervised,mandal2019out,xu2017transductive,xu2020transductive,zhuo2022zero}. Transductive learning performs inference over the entire video batch, rather than each video individually. Different from existing approaches, we use the test video distribution to improve the unseen action embeddings in the shared space of universal models by building on optimal transport. Moreover, our approach can switch between inductive and transductive settings, as only the positions of unseen action prototypes are updated.

\reb{\cite{wang2021zero} have also investigated optimal transport in the context of zero-shot learning. In their work, optimal transport is used to match distributions of generated and real features in their generative learning. \cite{wu2022data} consider optimal transport to align batches of unpaired images and textual descriptions for self-supervised learning with zero-shot learning as down-stream task. In our work, we outline a hyperspherical optimal transport to align distributions of unseen action labels to videos and objects in shared semantic spaces for zero-shot recognition in videos.}

\subsection{Zero-shot action localization}

Beyond recognition, a number of works have researched action localization in zero-shot settings. In the temporal domain, zero-shot localization has been investigated by aligning temporal proposals with label embeddings~\citep{zhang2020zstad, zhang2022tn}, by training models on trimmed seen actions followed by a knowledge transfer to unseen actions~\citep{jain2020actionbytes}, or by taking an open-set perspective~\citep{bao2022opental}.

In the spatio-temporal domain, \cite{jain2015objects2action} were the first to investigate zero-shot spatio-temporal action localization with a universal object perspective by computing object likelihoods for spatio-temporal proposals, followed by a semantic transfer to unseen actions. This direction has been expanded by incorporating knowledge about spatial relations between actors and objects \citep{mettes2017spatial} and by taking into account semantic priors about objects~\citep{mettes2021object}. We build upon these works and improve the ranking of action tubes from different videos through our universal prototype transport.

\cite{soomro2017unsupervised} have previously performed unsupervised spatio-temporal action localization, while \cite{yang2022less} have recently enabled a similar localization by performing a self-shot learning to automatically find relevant common videos from an unlabelled video pool to help the optimization. Different from these works, we seek to perform zero-shot spatio-temporal localization by assigning semantic labels to action tubes, rather than unsupervised or common action localization.

%% file: 4-method.tex
\section{Background}
%\psmm{To do: new background section on optimal transport.}
This paper builds upon optimal transport to improve zero-shot recognition. The background section provides the background for the general problem and the common task to be solved in optimal transport, following the formulation of \cite{peyre2019computational}. Optimal transport is the problem of moving one distribution of mass to another with minimal effort, with piles of dirt and multiple holes as a classical practical example.

More formally, optimal transport is a minimization problem over discrete measures, where a discrete measure is defined as:
\begin{equation}
\mu = \sum_{i=1}^{n} \mathbf{a}_i \delta_{x_{i}},
\end{equation}
with $\delta_{x_{i}}$ the Dirac position of the $i^{th}$ element and $\mathbf{a}_i \in \Sigma_n$ denotes probability vector and element of the probability simplex:
\begin{equation}
\Sigma_n = \{ \mathbf{a} \in \mathbb{R}^{n}_{+} : \sum_{i=1}^{n} \mathbf{a}_i = 1 \}.
\end{equation}
Optimal transport is concerned with finding an optimal assignment between two discrete measures. If we assume that two discrete measures are of equal size and want to find a one-to-one mapping between the elements of the two measures, we arrive at the Monge problem \citep{monge1781memoire}:
\begin{equation}
\min_{\sigma \in \text{Perm}(n)} \frac{1}{n} \sum_{i=1}^{n} \mathbf{C}_{i, \sigma(i)}
\end{equation}
with $\mathbf{C}_{i,j}$ a precomputed cost matrix between two discrete measures \reb{and Perm$(\cdot)$ the set of all possible permuations}. Here, we are interested in optimal assignment between discrete measures of different sizes and distributing mass from any point in one discrete measure to multiple points in the other discrete measure. We will therefore focus on the Kantorovich relaxation of the Monge problem \citep{kantorovich1942translocation}. In this relaxation, the permutation operation is replaced by a coupling matrix $\mathbf{P} \in \mathbb{R}^{n \times m}_{+}$, where $\mathbf{P}_{i,j}$ denotes the amount of mass that is distributed from point $i$ to point $j$. The minimization problem is in turn given as:
\begin{equation}
\mathcal{L}_{K}(\mathbf{a}_1, \mathbf{a}_2 | C) = \min_{\mathbf{P} \in \mathbf{U}(\mathbf{a}_1, \mathbf{a}_2)} \langle \mathbf{C}, \mathbf{P} \rangle = \sum_{ij} \mathbf{C}_{ij} \mathbf{P}_{ij},
\label{eq:kant}
\end{equation}
with $\mathbf{a}_1$ and $\mathbf{a}_2$ two discrete measures and $\mathbf{U}(\mathbf{a}_1, \mathbf{a}_2)$ the set of possible admissible couplings. With $\mathbf{a}_1$, $\mathbf{a}_2$, and $\mathbf{C}$ known, the goal is to find the optimal coupling matrix. We will extensively rely on optimal transport on the hypersphere and on the coupling matrix of the Kantorovich relaxation in the context of zero-shot actions. For a full foundation on optimal transport, we recommend the work of \cite{peyre2019computational}.

\begin{figure*}[t]
\centering
\includegraphics[width=\textwidth]{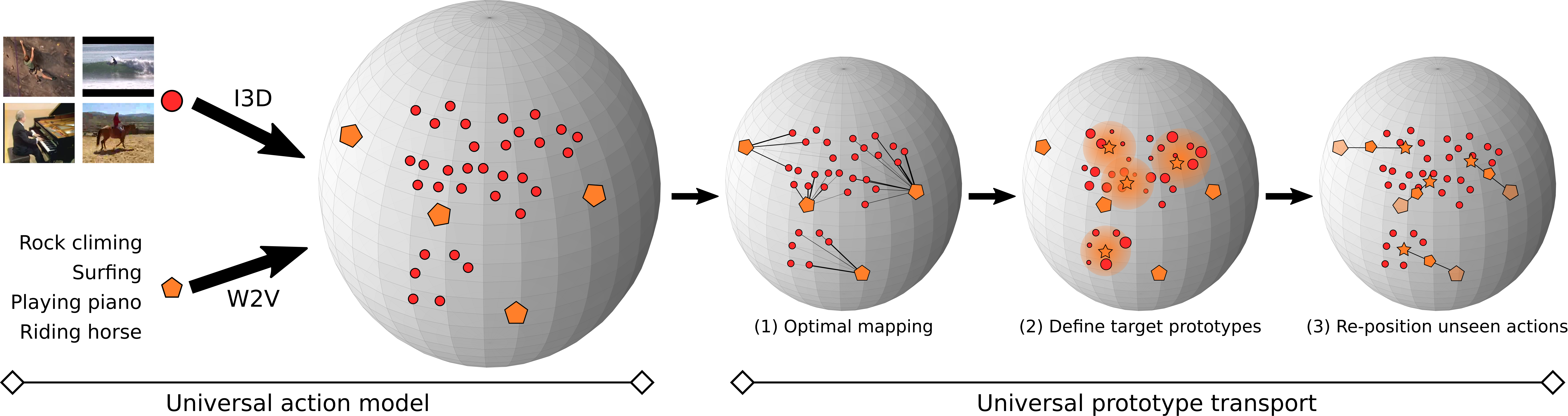}
\caption{\textbf{Overview of universal prototype transport.} First, we find an optimal mapping from unseen action prototypes to the projected test videos when building on universal action models. For universal object models, the test videos are replaced by object prototypes. Second, we define the target prototype for each unseen action as the weighted Fr\'echet mean over the transport couplings. Third, we re-position unseen action prototypes along the geodesic spanned by the original and target prototypes.}
\label{fig:method}
\end{figure*}

\section{Universal prototype transport}
%\psmm{To do: adjust method section based on background text and notation.}
For the problem of zero-shot action recognition, we are given a set of test videos $\mathcal{V}_u$ and a set of labels $L_u$ denoting actions which have not been observed during training. We seek to assign a label $l \in L_u$ to each test video. We start from two state-of-the-art universal learning directions in zero-shot action recognition, namely by transferring knowledge from large-scale seen actions and from objects.
%For both directions, we seek to overcome the biased and uneven label assignment of unseen actions during zero-shot inference.
Below, we first introduce transductive universal transport for the transfer from seen to unseen actions. Second, we extend our approach for re-positioning unseen action embeddings based on universal object models.

\subsection{Transporting universal action models}
Universal action models are centered around a semantic space that should be shared by both videos and action labels. This requires two transformation functions: a function $\omega$ that maps a label to a prototype in the semantic embedding space and a function $\phi$ that maps a video to the same embedding space. The function $\omega$ is given by a pre-trained word embedding model~\citep{mikolov2013distributed}, where embeddings are $\ell_2$-normalized and optimized with the cosine distance. The mapping function $\phi$ is learned on a set of training videos $\mathcal{V}_s$ with seen action labels $L_s$, with its loss given as:
\begin{equation}
\mathcal{L}_s = \sum_{v \in \mathcal{V}_s} - \frac{\langle \phi(v), \omega(l_v) \rangle}{||\phi(v)|| \cdot ||\omega(l_v)||},
\end{equation}
with $l_v \in L_s$ the action label for video $v$. Once such a network is optimized for a set of seen actions with training videos, zero-shot learning can be enabled for a test video simply through a nearest neighbor search with a set of unseen action prototypes in the shared semantic space. This paper argues that that since $\phi$ is trained on actions not used during inference, the projected videos and unseen action prototypes are not well aligned. We propose to improve zero-shot action recognition by re-positioning unseen action prototypes with optimal transport. 

The main idea here is to find an optimal mapping between the set of unseen action prototypes and the set of projected test videos in the shared semantic space. We then want to utilize the optimal mapping to improve the location of unseen action prototypes in the shared space, to improve the zero-shot nearest neighbor inference. Figure~\ref{fig:method} provides an overview of our approach. To be able to find the optimal mapping between unseen actions and test videos, we first need to redefine them as discrete measures in order to solve the corresponding optimal transport problem. For the unseen actions, the definition is given as:
\begin{definition}
\emph{(Actions as a discrete measure).} The set of unseen actions is represented as a measure as:
\begin{equation}
\mu_u = \sum_{l_u \in L_u} w_{l_u} \delta_{\omega(l_u)},
\end{equation}
where $w_{l_u} \in \mathbf{w}_u$ denotes the set of weights for the action, such that $\mathbf{w}_u \in \Sigma_{|L_u|}$ is on the probability simplex, and $l_u$ denotes the label of unseen action $u$.
\label{def:action}
\end{definition}
The unseen actions are given by a weighted combination of their word embeddings in the shared semantic space. The definition of the projected test videos is given as:
\begin{definition}
\emph{(Videos as a discrete measure).} The set of projected test videos is given as a measure as:
\begin{equation}
\mu_v = \sum_{c \in C^k(\mathcal{V}_u)} w_c \delta_c, \quad \quad c = \frac{1}{|a(\mathcal{V}_u; c)|} \sum_{v \in a(\mathcal{V}_u; c)} \phi(v),
\end{equation}
with $w_c \in \mathbf{w}_c \in \Sigma_k$ akin to Definition \ref{def:action}, where $C^k$ denotes a $k$-component cluster aggregation over the set of videos, and where $a(\mathcal{V}_u; c)$ denotes the set of videos assigned to cluster $c$.
\label{def:video}
\end{definition}
Rather than using all test videos as individual points in the discrete measure, we first cluster the test videos and define the measure over the cluster centers. The idea behind Definition \ref{def:video} is to make the discrete measure robust to outliers and to increase the focus on high-density regions of projected videos.

With the unseen action labels and videos defined as discrete measures in the same space, we are able to compute an optimal transport mapping between the two. By operating over the entire distribution of test videos rather than performing inference for each video independently, we view this as a transductive form of optimal transport. We seek to obtain a coupling matrix $\mathbf{P}^u$ with the minimization objection given in Equation~\ref{eq:kant}. To that end, let $\mathbf{C}^u$ denote the required cost matrix, with $C_{ij}^u$ defined as the cosine distance between unseen action $i$ and video cluster $j$. Then the hyperspherical optimal transport from $\mu_u$ to $\mu_v$ is as:
\begin{equation}
\mathcal{L}_{K}(\mathbf{w}_u, \mathbf{w}_v | \mathbf{C}^u) = \min_{\mathbf{P}^u \in \mathbf{U}(\mathbf{w}_u, \mathbf{w}_v)} \langle \mathbf{C}^u, \mathbf{P}^u \rangle = \sum_{ij} C^u_{ij} P^u_{ij},
\label{eq:emd1}
\end{equation}
\reb{with $P^u_{ij}$ a single coupling value}, where the minimization is solved using the Lagrangian approach of~\cite{bonneel2011displacement}. This results in a coupling matrix $\mathbf{P}^u$. When working with universal action models, we set the weights $\mathbf{w}_u$ and $\mathbf{w}_v$ uniformly. The overall step of finding an optimal mapping from unseen actions to test videos is shown in Figure~\ref{fig:method} as step (1) in the universal prototype transport.

Given the optimal transport mapping, we propose to condense the corresponding coupling into a single target prototype per unseen action. Since the semantic space on which we operate is a hypersphere, we define the target prototype for unseen action $i$ as the weighted Fr\'{e}chet mean~\citep{lou2020differentiating,miolane2020geomstats} based on normalized coupling values:
\begin{equation}
\omega^{\text{target}}(l_i) = \argmin_{s \in \mathbb{S}^{d-1}} \sum_{j=1}^{k} \hat{P}_{ij} d(c_j, s)^2, \quad \mathbf{\hat{P}} = \mathbf{P}^u / ||\mathbf{P}^u||_1.
%\sum_{j=1}^{k} \hat{P}_{ij} = 1,
\label{eq:interpolate1}
\end{equation}
with $d$ the cosine similarity \reb{and $s$ the obtained mean}. Determining the target prototype of each unseen action is visualized in Figure~\ref{fig:method} as step (2).
\nreb{We opt for a hyperspherical optimal transport formulation because we rely on word embeddings for actions and objects, which are hyperspherical in nature as they are optimized with cosine distances \citep{mikolov2013distributed} and they are state-of-the-art for zero-shot action recognition \citep{brattoli2020rethinking, pu2022alignment, zhuo2022zero}.}
%\reb{We opt for a hyperspherical optimal transport formulation, as state-of-the-art zero-shot action recognition methods operate on word embeddings for actions and objects, which operate on hyperspherical manifolds.}
The target provides a new prototype in embedding space for each unseen action, guided by the distribution of mapped and clustered test videos. While we can directly use the new embeddings for inference, we want to avoid big changes in embedding space since that relates with losing the original semantic interpretation of the action. We therefore dictate that the final prototype of each unseen action is positioned along the geodesic spanned by the original and target prototypes, modelled through spherical interpolation:
\begin{equation}
\begin{split}
\omega^{\star}(l) = & \frac{\sin [\lambda \Omega]}{\sin \Omega} \omega(l) + \frac{\sin [(1-\lambda) \Omega]}{\sin \Omega} \omega^{\text{target}}(l),\\
\cos \Omega = & \langle \omega(l), \omega^{\text{target}}(l) \rangle,
\end{split}
\label{eq:interpolate2}
\end{equation}
where $\lambda$ denotes the interpolation ratio between the extremes. In this manner, we move each unseen action towards its proposed target prototype, with the interpolation acting as a regularization pulling the action towards the original semantic prototype, visualized as step (3) in Figure~\ref{fig:method}. Zero-shot inference is performed the same as in existing universal action models, by means of a nearest neighbour search between video $v$ and each unseen action label $l$ as
\begin{equation}
s_{\text{action}}(l | v) = \frac{\langle \phi(v), \omega^{\star}(l) \rangle}{||\phi(v)|| \cdot ||\omega^{\star}(l)||},
\label{eq:score1}
\end{equation}
after which the action label with the highest similarity is selected.

\subsection{Transporting universal object models}
Universal object models for zero-shot action recognition suffer from the same bias in the assignment of unseen action labels to test videos. We therefore also seek to transport unseen action prototypes in this setting. We start by redefining objects as discrete measures to make them suitable in the context of optimal transport:
\begin{definition}
\emph{(Objects as a discrete measure).} The set of objects are given as a measure as :
\begin{equation}
    \mu_o = \sum_{o \in \mathcal{O}_s} w_{o} \delta_{\omega(o)}, \quad \mathcal{O}_s = \{ o \in \mathcal{O} \ | \ \max_{v \in \mathcal{V}_u} p(o|v) \geq \tau \},
\end{equation}
with $w_o \in \mathbf{o}_c$ and where $p(o|v)$ denotes the likelihood of object $o$ occurring in video $v$.
\end{definition}
Unique in this definition, the discrete measure for objects is based on a subset $\mathcal{O}_s \in \mathcal{O}$, \ie we define a degenerate distribution over objects. The subset is determined by again taking a transductive view; we exclude any object which does not have a likelihood over a low threshold $\tau$ in \emph{any} test video. The idea behind this is simple: we want to avoid a bias in the optimal transport to objects which do not actually occur in videos.

In the universal object context, the optimal transport mapping is now given between the unseen action measure and the object measure. Moreover, we set non-uniform weights for both the actions and objects. The objects are weighted according to their transductive maximum score,
\begin{equation}
w_o = \max_{v \in \mathcal{V}_u} p(o|v) / \mathcal{Z}_o,
\end{equation}
with $\mathcal{Z}_o$ a normalization constant over all objects in $\mathcal{O}_s$. The unseen action weights are given as
\begin{equation}
w_a = (1 - ((\max_{o \in \mathcal{O}_s} \langle \omega(a), \omega(o) \rangle / 2) + 0.5)) / \mathcal{Z}_a,
\end{equation}
with $\mathcal{Z}_a$ a normalization constant over all actions. The intuition behind the object weights is to focus the attention of the transductive universal transport on objects with a higher visual likelihood. The weights for the unseen actions are given as the inverse over the maximum word embedding similarity with respect to the objects, under the notion that actions without obvious relations to objects should have a more prominent spot in the transport coupling. With the optimal transport computed between unseen actions and objects, action prototypes are again interpolated following Equations~\ref{eq:interpolate1} and~\ref{eq:interpolate2}, with the updated prototype for action label $l$ now denoted as $\omega^{\ddagger}(l)$.

For the final zero-shot action inference from objects, we follow the same setup as current object-based approaches, where the score for each action label $l$ in video $v$ is determined based on the top relevant objects for that action~\citep{jain2015objects2action,mettes2021object}:
\begin{equation}
s_{\text{object}}(l | v) = \sum_{o \in \mathcal{O}_l} p(o | v) \cdot \frac{\langle \omega(o), \omega^{\ddagger}(l) \rangle}{||\omega(o)|| \cdot ||\omega^{\ddagger}(l)||}.
\label{eq:score2}
\end{equation}
with $\mathcal{O}_l$ the set of most semantically similar objects for action label $l$. Finally, the transductive action-based and object-based scores from respectively Equations~\ref{eq:score1} and~\ref{eq:score2} can also be fused as $s_{\text{fusion}}(l | v) = \epsilon \cdot s_{\text{action}}(l | v) + (1 - \epsilon) \cdot s_{\text{object}}(l | v)$.

\reb{Summarized, our approach formulated for universal object models differs in three ways from its formulation to universal action models: the mapping is performed towards object embeddings rather than video embeddings, the object measure only includes objects that actually occur in any of the test videos, and the object measure is weighted based on video likelihood, where as the video measure is unweighted.}

%% file: 5-experiments.tex
\section{Experimental setup}

\subsection{Datasets}

\textbf{Source datasets.}
We employ two source datasets for universal representation learning, namely Kinetics-700 for actions and ImageNet for objects. For \textbf{Kinetics-700}, we follow~\cite{brattoli2020rethinking} and use a subset with 664 action categories to avoid any overlap with action categories in datasets used for zero-shot action recognition. For \textbf{ImageNet}, we follow~\cite{mettes2021object} and use the reorganized variant containing 12,988 object categories~\citep{mettes2020shuffled}.
\\\\
\textbf{Target datasets.}
The classification evaluation is performed on the two datasets used most often in zero-shot action recognition: UCF-101 and HMDB51. The \textbf{UCF-101} dataset consists of 13,320 videos covering 101 action categories. Next to 101-way zero-shot evaluation, we also evaluate on settings with 20 and 50 test actions. For these settings, we rerun our approach on 10 runs with randomly selected actions and we report the mean and standard deviation over the runs. We note that in the 20- and 50-way zero-shot recognition, we do not use the other actions for training, they are simply not used in our approach. The \textbf{HMDB51} dataset consists of 6,766 videos covering 51 action categories. Next to 51-way evaluation, we also investigate 10- and 25-way zero-shot recognition.

\begin{figure*}[t]
\centering
\begin{subfigure}{0.495\textwidth}
\includegraphics[width=\textwidth]{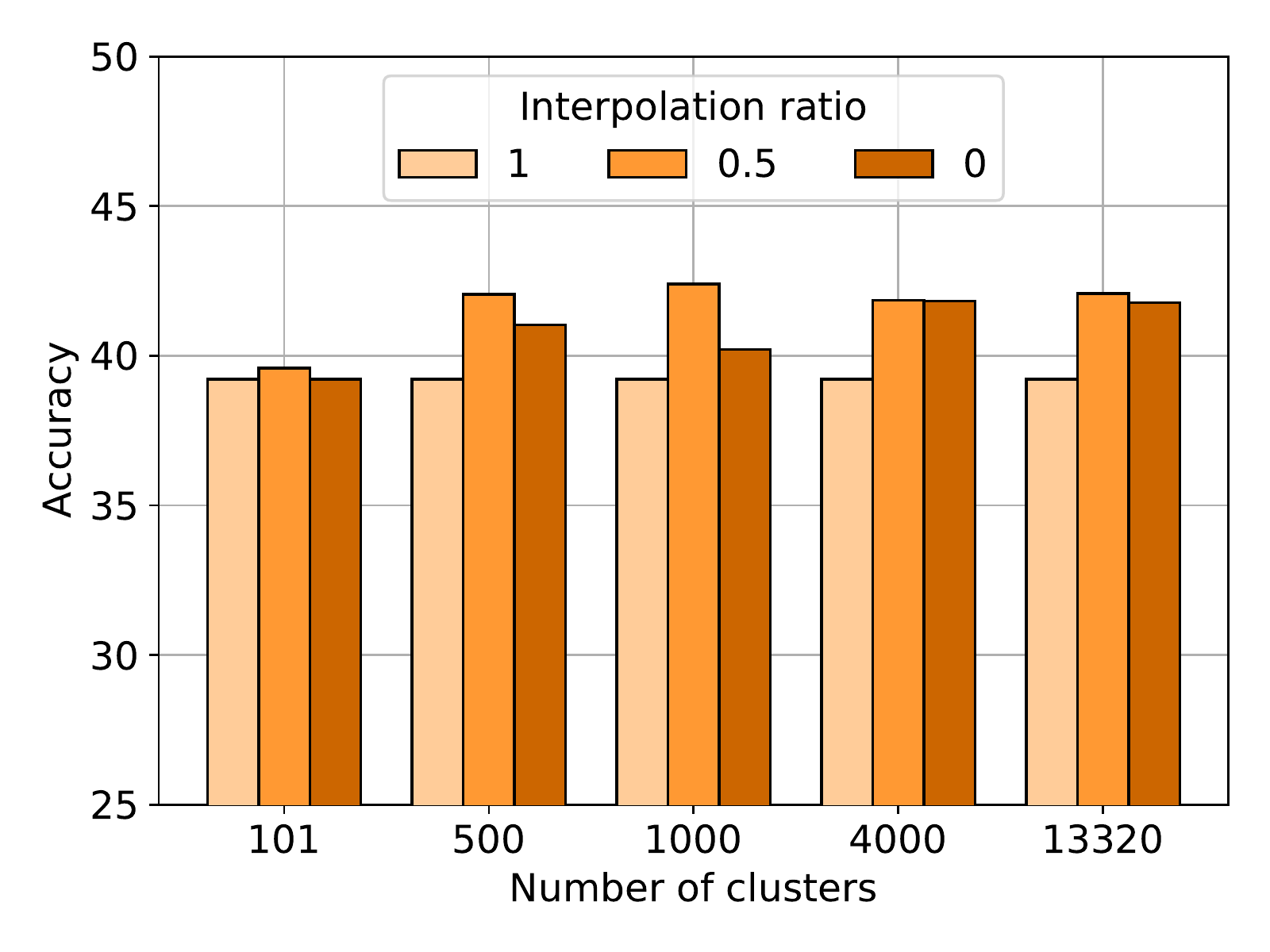}
\caption{Cluster ablation.}
\label{fig:abl1a}
\end{subfigure}
\begin{subfigure}{0.495\textwidth}
\includegraphics[width=\textwidth]{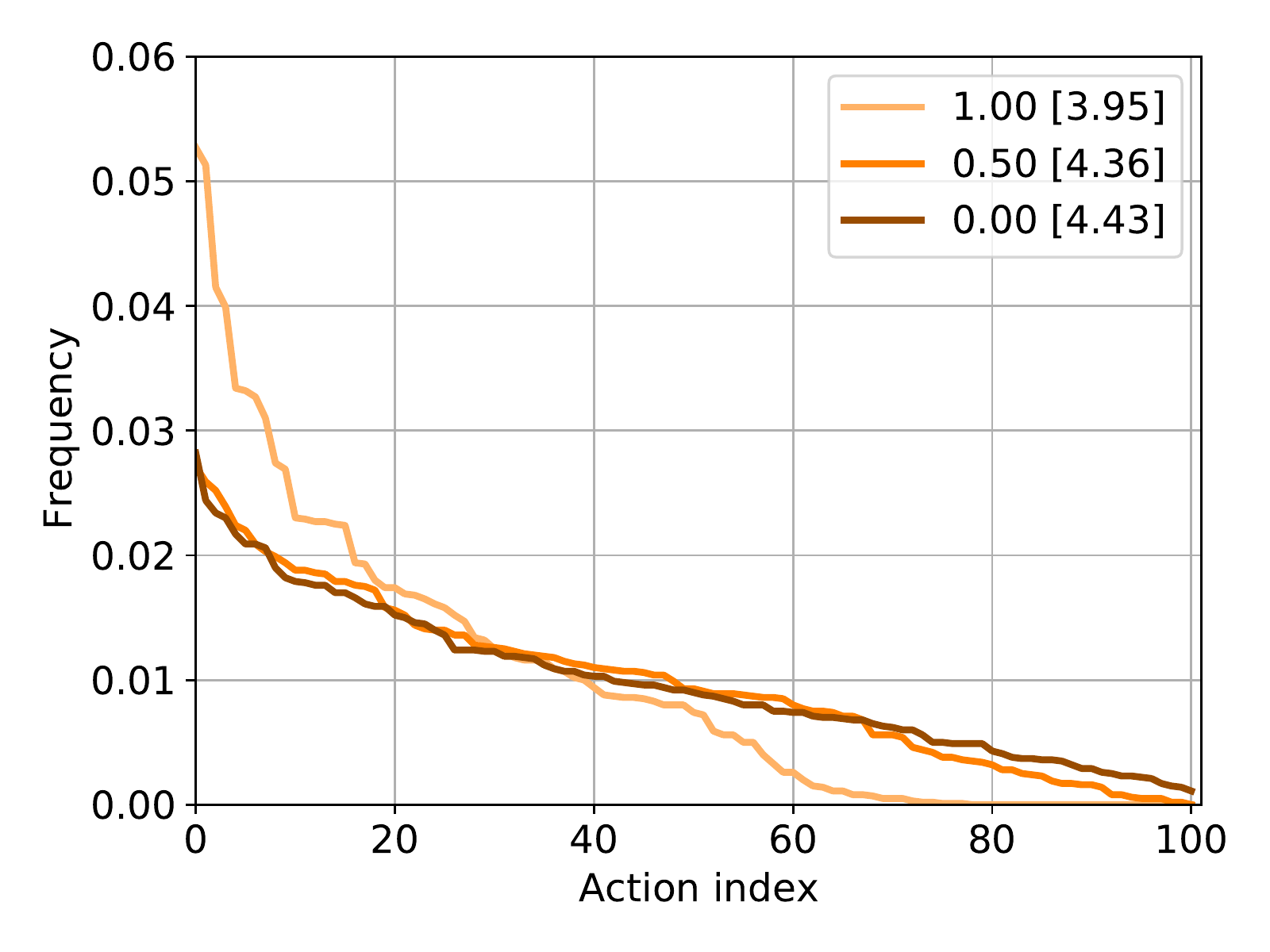}
\caption{Distribution analysis (1,000 clusters).}
\label{fig:abl1b}
\end{subfigure}
\caption{\textbf{Evaluating universal prototype transport from seen actions} on UCF-101. Left: The effect of the number of clusters and the interpolation ratio on the recognition performance. An interpolation rate of 1 denotes the baseline with the original unseen action prototypes. We find that re-positioning the prototypes directly boosts zero-shot performance given sufficient clusters, with a further boost by interpolating between the original and target prototypes, see the highest overall score for 1,000 clusters and interpolation ratio 0.5. Right: Intuition behind our improved results. Using the original unseen action prototypes results in large biases during zero-shot inference. With our approach, this imbalance is reduced, as indicated by the more even action distributions in the plot and the corresponding higher entropy scores in brackets in the legend.}
\label{fig:abl1}
\end{figure*}

We also investigate the potential of our approach for zero-shot spatio-temporal action localization on UCF Sports and J-HMDB. \textbf{UCF Sports} consists of 150 videos with 10 actions and \textbf{J-HMDB} consists of 928 videos with 21 actions. For the evaluation, we follow~\cite{jain2015objects2action} and report results with the AUC metric across five overlap thresholds.

\subsection{Implementation details}
We consider two universal action networks $\phi$.
First, we employ the R(2+1)D network~\citep{tran2018closer} as given by~\cite{brattoli2020rethinking}, pre-trained on 664 Kinetics categories. For each video, we obtain its corresponding video embedding by randomly selecting a 16 frame shot and passing the shot through $\phi$. For a fair comparison to~\cite{brattoli2020rethinking}, we also use the same word embedding $\omega$, namely a word2vec model~\citep{mikolov2013efficient}, resulting in a 300-dimensional representation per word. For any action or object with more than one word, the word representations are averaged. Second, we employ the network of~\cite{pu2022alignment} with 25 splits per video, which is based on the same word embedding and Kinetics splits as the first action model. For both approaches, we rely on author-provided code to obtain action and video embeddings. For the universal object model, the object scores in a video are obtained following the protocol of~\cite{mettes2021object}, where two frames per second are sampled, each fed to the pre-trained ImageNet model, and with the object probabilities averaged over the sampled frames.

For the clustering of the video embeddings, we use k-means clustering along with $\ell_2$-normalization, akin to~\cite{banerjee2005clustering}. For the optimal transport, we employ the Lagrangian approach of~\cite{bonneel2011displacement} as implemented in~\citep{flamary2021pot}. \reb{Specifically, we set the cosine distance matrix as loss matrix, run for a maximum of 100,000 iterations if there has been no convergence and with the dual potential centered in the optimization.} Unless specified otherwise, accuracy denotes the top 1 accuracy. Lastly for spatio-temporal localization, we start from the tubes made available by~\cite{mettes2021object}. To each tube, we add the corresponding video-level action scores from our approach to improve the ranking of the action tubes over the entire dataset. \nreb{The universal transport takes roughly 36 seconds CPU time for 13,320 videos, 101 actions, and 1000 clusters on UCF-101 on an Intel Xeon CPU. Once the action prototypes are re-positioned, no additional computational effort is required for zero-shot inference.} All code will be made publicly available.

\section{Experimental results}

%
% START
%
We focus on five experiments: (i) evaluations on universal action models; (ii) evaluations on universal object models; (iii) integrating and fusing our approach with recent methods; (iv) state-of-the-art comparison for zero-shot action recognition and zero-shot spatio-temporal action localization; (v) qualitative analyses.

%
%%%%% EXP 1
%
\subsection{Universal transport from seen actions}

\noindent
\textbf{Setup.} For the first experiment, we evaluate on UCF-101 using all 101 actions for classification. We investigate the two variables that come with our approach in the context of universal action models, namely the granularity of the cluster aggregation over all test videos and the interpolation ratio between the original and target prototypes of the unseen actions. We use the universal action model of \cite{brattoli2020rethinking} throughout this experiment.
\\\\
\textbf{Results.} The results for five cluster sizes and three interpolation ratios are shown in Figure~\ref{fig:abl1a}. An interpolation rate of 1 denotes the baseline using only the original action prototypes and 0 denotes the setting using the target prototypes. With the original semantic prototypes, we obtain an accuracy of 39.2\%. Using the target prototypes directly boosts the classification accuracy when using sufficiently many clusters. Using only few clusters results in a coarse approximation of the distribution of test videos, which leads to lower performance. The best performance is obtained by positioning the unseen actions halfway along the geodesic between the original and target embeddings. With 1,000 clusters the accuracy becomes 42.4\%, compared to 40.1\% when using the target prototypes directly. We will use 1,000 clusters and an interpolation ratio of 0.5 for all other experiments involving universal action models.
\\\\
\textbf{Analysis.} An explanation for our obtained improvements is shown in Figure~\ref{fig:abl1b}. We show the distributions of selected actions across all three interpolation ratios when using 1,000 clusters. With the original unseen action embeddings, this distribution is highly uneven, with 23\% of the actions never being selected, naturally leading to zero accuracy for these actions. With our approach, the distributions become more uniform, highlighting the bias reduction. This is also reflected in the entropy of the action selection distributions in the top right of~\ref{fig:abl1b}, which increases when employing universal prototype transport, confirming that the distribution becomes more uniform.

\begin{figure}[t]
\centering
\includegraphics[width=\linewidth]{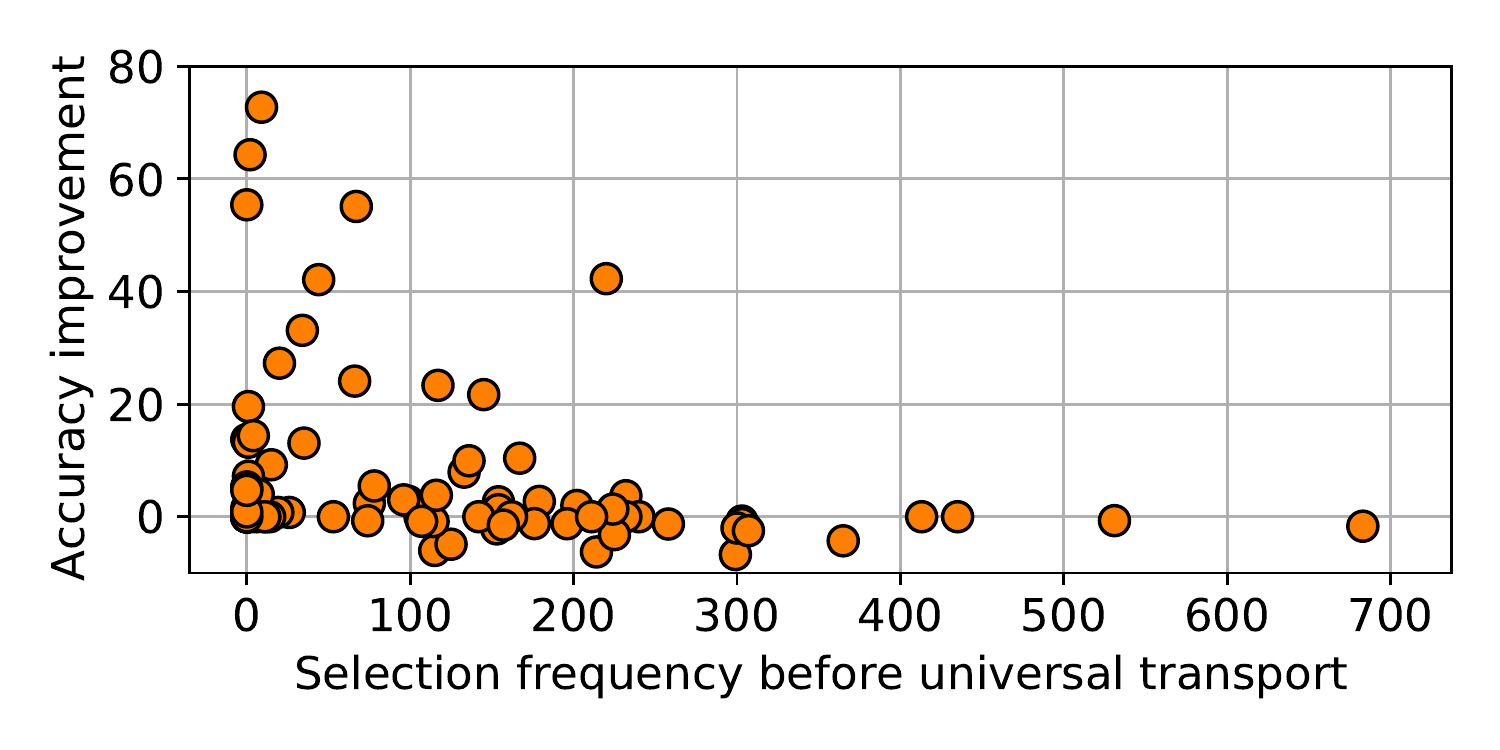}
\caption{\textbf{Per-class improvements} as a function of the number of times an action is selected prior to universal prototype transport on UCF-101. Our approach improves especially those classes that are rarely selected in a stand-alone universal action model.}
\label{fig:perclass}
\end{figure}

\begin{figure}[t]
\centering
\begin{subfigure}{0.49\linewidth}
\includegraphics[width=\linewidth]{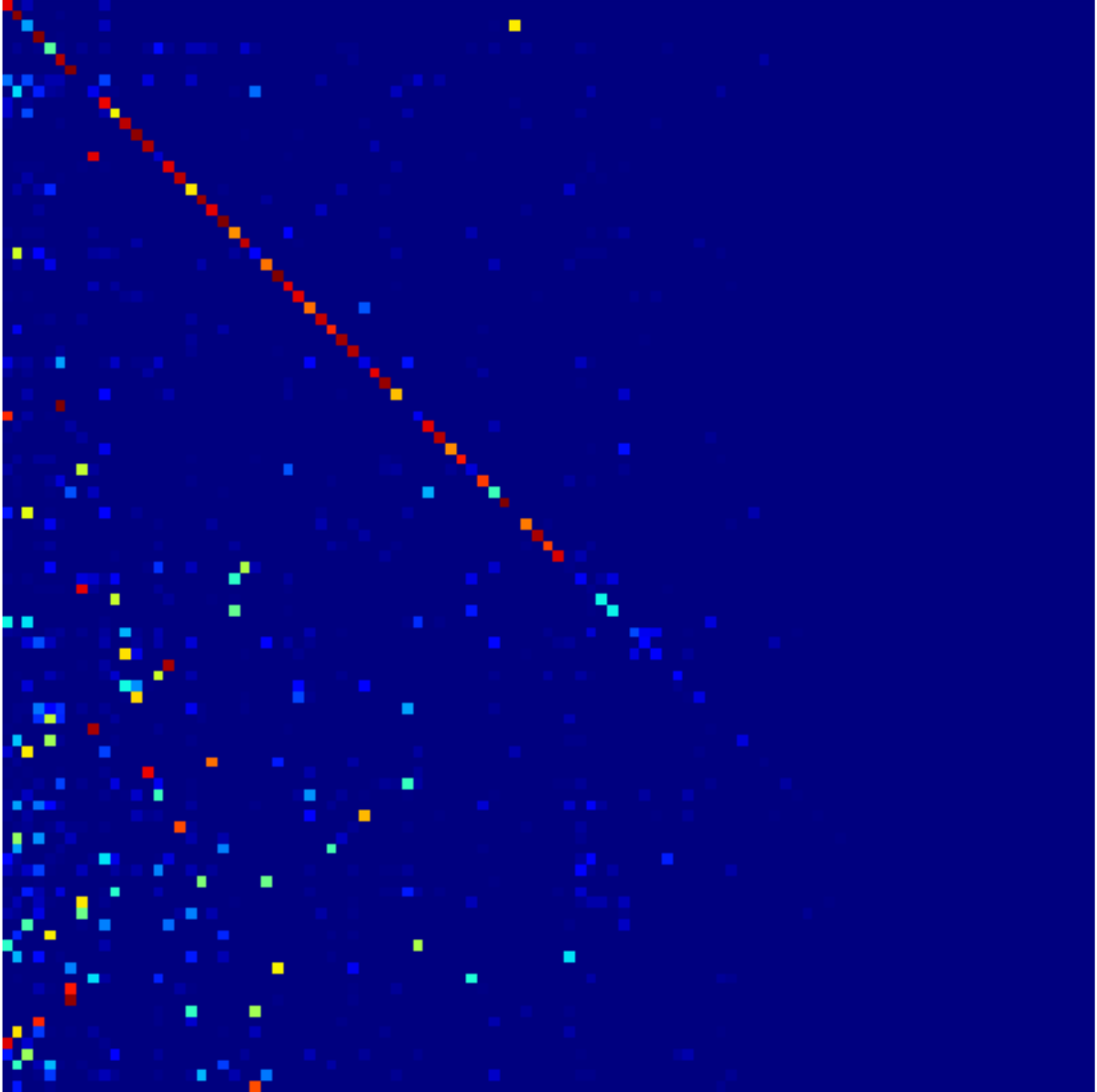}
\caption{\citep{brattoli2020rethinking}.}
\label{fig:confmata}
\end{subfigure}
\begin{subfigure}{0.49\linewidth}
\includegraphics[width=\linewidth]{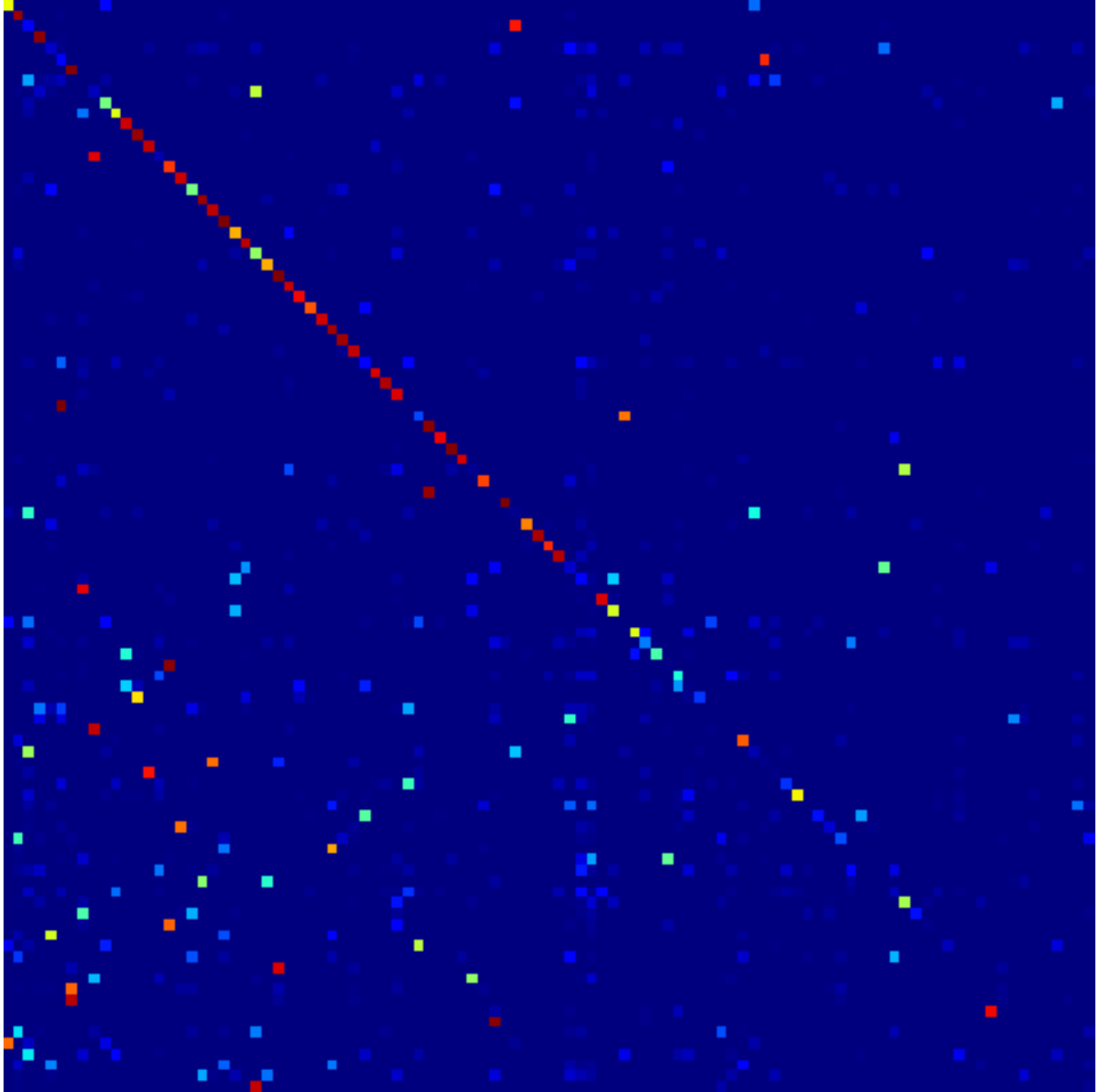}
\caption{This paper.}
\label{fig:confmatb}
\end{subfigure}
\caption{\reb{\textbf{Confusion matrices on UCF-101} before (left) and after (right) performing universal prototype transport on the model by \cite{brattoli2020rethinking}, sorted by selection frequency in the base model. For the baseline, the entire right side of the matrix is dark blue, since those actions are never selected by any test video. After performing our approach, the confusion matrix is more uniform and highlights a better performance for actions ignored by the baseline.}}
\label{fig:confmat}
\end{figure}

%We furthermore look at the class-level results in Figure~\ref{fig:perclass}. The figure shows that our approach improves especially those classes that were not frequently occurring in a stand-alone universal model. This result highlights that our improvements are due to a better alignment between unseen action prototypes and projected test videos in semantic space. We conclude that universal prototype transport improves zero-shot action recognition for universal action models.
\reb{In Figure~\ref{fig:perclass}, we show the per-class performance gains as a function of selection frequency before our transport. The figure shows that our approach improves especially those classes that were not frequently occurring in a stand-alone universal model. This result highlights that our improvements are due to a better alignment between unseen action prototypes and projected test videos in semantic space. Examples of most improved actions include sky diving (accuracy from 8.2\% to 80.9\%), hammering (from <0.1\% to 65.0\%), and cricket bowling (from 0.0\% to 55.4\%). Figure~\ref{fig:confmat} provides dives deeper into the observation by highlighting the improved alignment based on the entire confusion matrix.}

\nnreb{As an extra test, we investigate the effect of class imbalance in the test set for universal prototype transport. Following the long-tailed literature \citep{cui2019class}, we sample UCF-101 with exponential decays of factors 0.1 and 0.01. The average per-class accuracy on standard UCF-101 is 42.0\% and remains stable (42.4\% at imbalance ratio 0.1 and 41.6\% at imbalance ratio 0.01), highlighting that our approach is stable to test-time class imbalance.}

%
%%%%% EXP 2
%
\subsection{Universal transport from objects}

\noindent
\textbf{Setup.}
Second, we investigate our approach on universal object models. We again use UCF-101 with all 101 actions for evaluation, with the interpolation ratio fixed to 0.5. We evaluate three threshold levels that come with the definition of objects as discrete measure, along with the universal object model itself and a vanilla uniformly-weighted optimal transport using all objects as baselines.
\\\\
\textbf{Results.} In Table~\ref{tab:abl2-1}, we show the zero-shot action results for our approach when maintaining the top 2,500, 1,000, and 500 objects according to their transductive maximum likelihoods over all test videos. We first find that using a baseline optimal transport approach akin to the setup for seen actions provides only a marginal boost from 29.9\% to 30.1\%. In contrast, using the proposed weights for the unseen actions and objects, combined with a filtering of objects never present in a test video, provides a boost to 31.6\% with the top 1,000 objects. We find that as long as the threshold is not set too strictly (\eg keeping 1,000 objects or more) provides stable zero-shot results.

\begin{table}[t]
\centering
\begin{tabular}{l @{\hskip 4\tabcolsep} c @{\hskip 4\tabcolsep} c}
\toprule
\multicolumn{3}{c}{\textbf{UCF-101}}\\
 & \# objects & Accuracy\\
\midrule
Baseline & - & 29.9\\
Baseline transport & 12,988 & 30.1\\
\midrule
Proposed transport & 2,500 & 31.4\\
Proposed transport & 1,000 & \textbf{31.6}\\
Proposed transport & 500 & 30.2\\
\bottomrule
\end{tabular}
\caption{\textbf{Evaluating universal prototype transport from objects} on UCF-101. Our proposed approach also enhances universal object-based approaches for zero-shot action recognition, especially when incorporating object filtering.}
\label{tab:abl2-1}
\end{table}
\begin{table}[t]
\centering
\begin{tabular}{c @{\hskip 4\tabcolsep} c @{\hskip 4\tabcolsep} c}
\toprule
\multicolumn{3}{c}{\textbf{UCF-101}}\\
\multicolumn{2}{c}{Weighting} & Accuracy\\
Actions & Objects &\\
\midrule
uniform & uniform & 29.8\\
inverse & uniform & 30.4\\
uniform & transductive & 30.6\\
inverse & transductive & \textbf{31.6}\\
\bottomrule
\end{tabular}
\caption{\textbf{Proposed versus uniform weighting} between unseen actions and objects. Focusing on unseen actions with low semantic relation to any object (inverse) and on objects objects that are also observed in all test videos (transductive) improve the prototype transport from objects for zero-shot action recognition.}
\label{tab:abl2-2}
\end{table}

In Table~\ref{tab:abl2-2}, we show that the proposed weighting matters. For this Table we keep the top 1,000 objects and investigate all four combinations of uniform and proposed weighting. With uniform weights for both unseen actions and objects, the results are similar to the baseline object-based setup, while the results improve when incorporating either or both of the weight vectors to the proposed setup. We conclude that in the universal object-based model for zero-shot action recognition, the proposed transport is also beneficial.

\begin{figure*}[t]
\centering
\begin{subfigure}{0.495\textwidth}
\includegraphics[width=\textwidth]{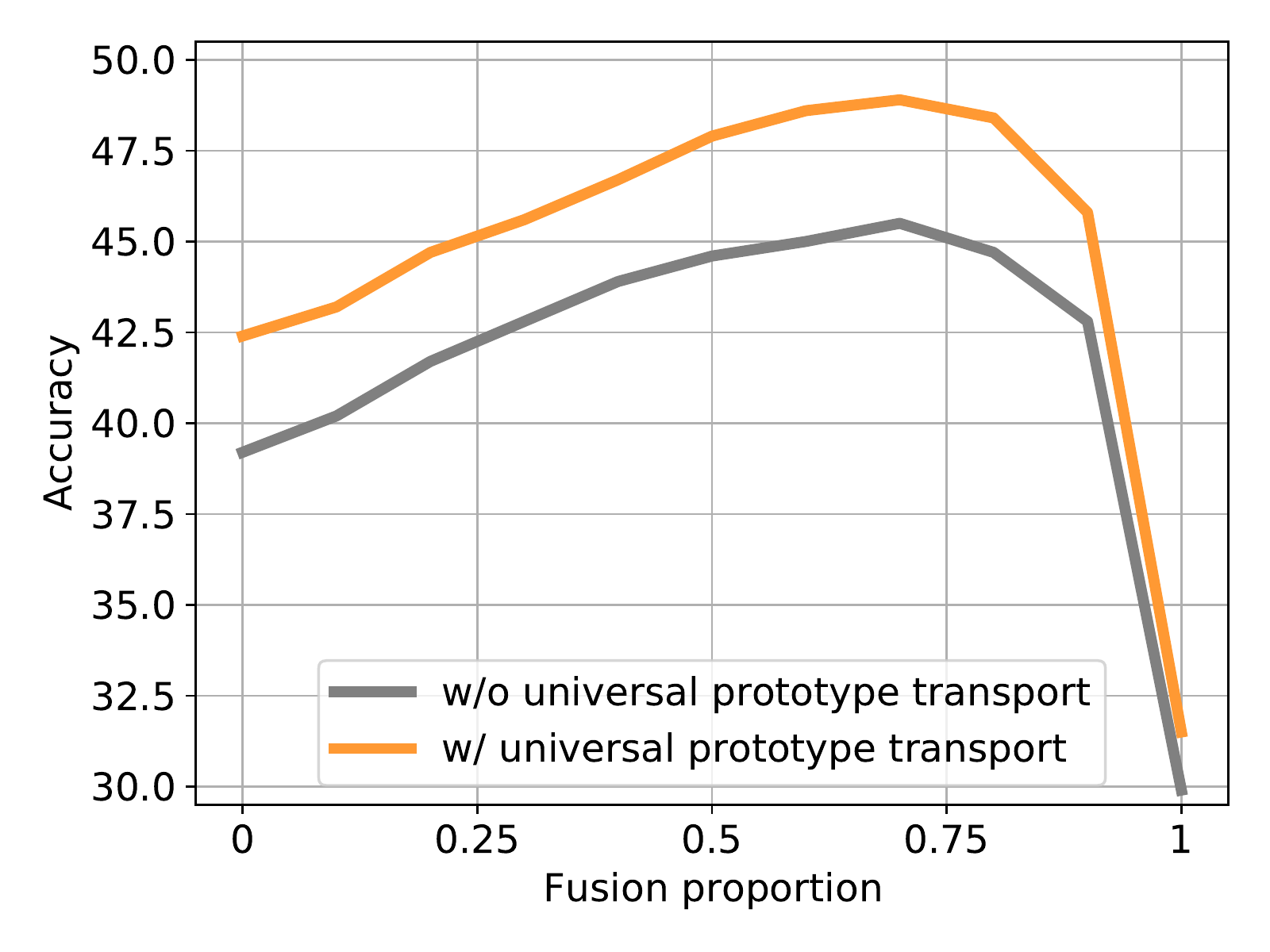}
\caption{UCF-101.}
\label{fig:fusiona}
\end{subfigure}
\begin{subfigure}{0.495\textwidth}
\includegraphics[width=\textwidth]{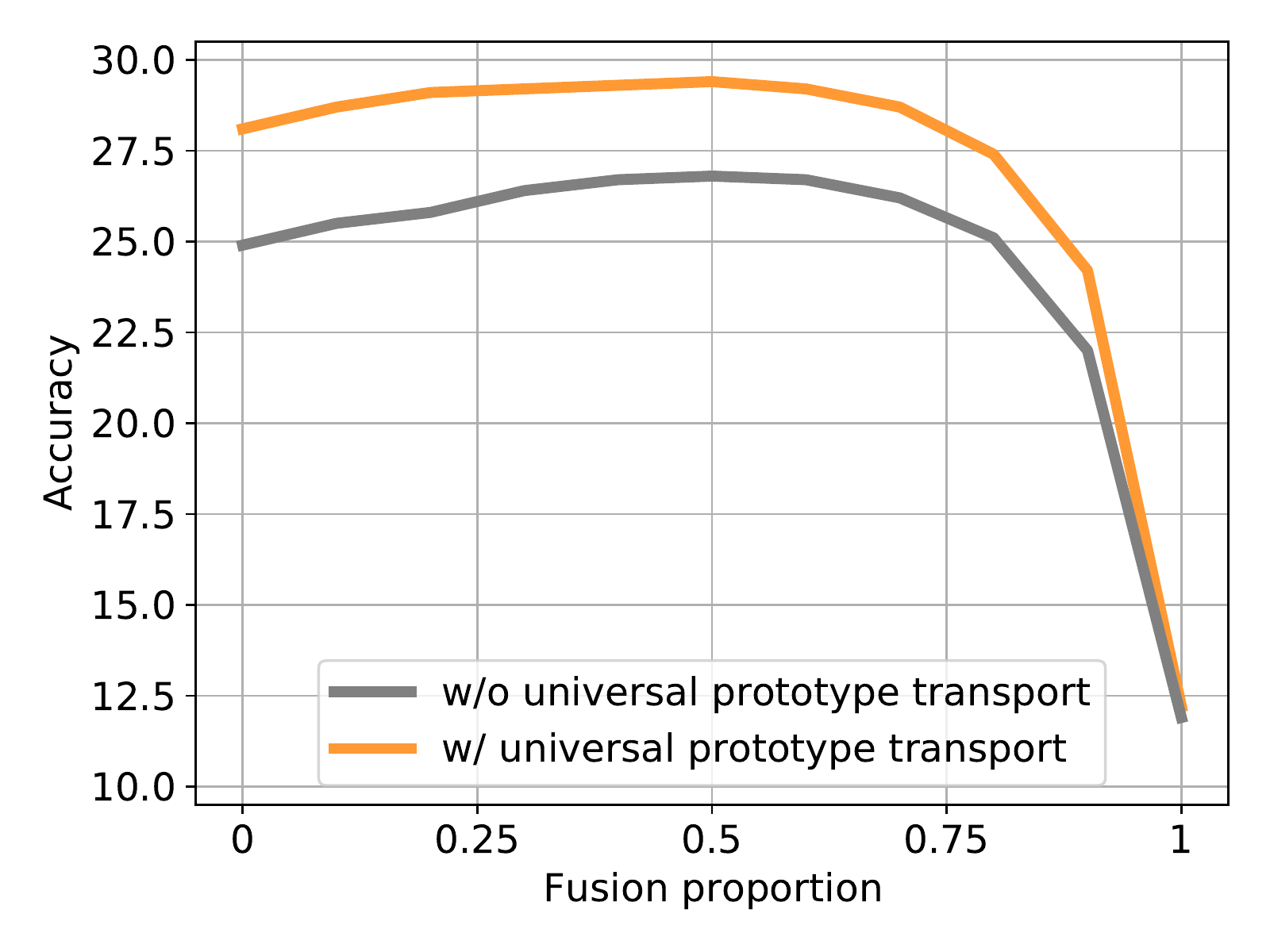}
\caption{HMDB51.}
\label{fig:fusionb}
\end{subfigure}
\caption{\textbf{Fusing action and object information} for zero-shot recognition on UCF-101 and HMDB51. Combining universal action and object information benefits zero-shot recognition, with universal prototype transport preferred across all fusion proportion between both sources.}
\label{fig:fusion}
\end{figure*}

%
%%%%% EXP 3
%
\subsection{Transporting multiple universal models}

\noindent
\textbf{Setup.}
Third, we investigate how universal prototype transport operates across and in combination with multiple universal models. We focus on two results: providing an overview of our approach on multiple state-of-the-art universal models and combining action and object models with universal transport.
\\\\
\textbf{Results.} In Table~\ref{tab:addsota}, we provide an overview of our approach on top of recent universal action and object models. For the universal action models of \cite{brattoli2020rethinking} and \cite{pu2022alignment}, we use the pre-trained models provided by the authors to compute video embeddings and add our universal prototype transport on top. For the universal object model of~\cite{mettes2021object}, we take the author-provided pre-trained object network and directly use the conventional object-to-action formulation of Equation~\ref{eq:score2} to obtain zero-shot predictions. We find that across multiple models, the results of both the top 1 and top 5 accuracy are improved. For action models, the top 5 improvements are even higher than the top 1 improvements, highlighting the better overall alignment between unseen actions and test videos. We conclude form this Table that our approach is generic and can be plugged in new methods to obtain better zero-shot results.

\begin{table}[t]
\centering
\begin{tabular}{l c cc}
\toprule
 & & \multicolumn{2}{c}{\textbf{UCF-101}}\\
 & & top-1 & top-5\\
\midrule
\cite{brattoli2020rethinking} & action & 39.2 & 60.8\\
\rowcolor{Gray}
\textbf{+ This paper} & & \textbf{42.4} & \textbf{69.2}\\
\midrule
\cite{pu2022alignment} & action & 46.3 & 72.9\\
\rowcolor{Gray}
\textbf{+ This paper} & & \textbf{47.0} & \textbf{74.4}\\
\midrule
\cite{mettes2021object} & object & 29.9 & 54.6\\
\rowcolor{Gray}
\textbf{+ This paper} & & \textbf{31.6} & \textbf{54.7}\\
\bottomrule
\end{tabular}
%\caption{\textbf{Incorporating universal prototype transport} in state-of-the-art action and object models on UCF-101. Incorporating our approach on top of recent methods improves both top 1 and top 5 accuracy for zero-shot action recognition.}
\caption{\nreb{\textbf{Universal prototype transport} on top of several inductive action and object models on UCF-101. We show here that our transductive approach is general in nature and can be plugged on top of different universal models, improving both top 1 and top 5 accuracy.}}
%\caption{\reb{\textbf{Incorporating universal prototype transport} in state-of-the-art inductive action and object models on UCF-101. Incorporating our transductive approach on top of recent methods improves both top 1 and top 5 accuracy for zero-shot action recognition.}}
\label{tab:addsota}
\end{table}

Beyond individual universal models, we also investigate the potential of combining action with object models for zero-shot action recognition. Intuitively, both types of methods bring different perspectives and rely on different sources for generalizing to actions without training examples. Hence their predictions can be of a complementary nature. In Figure~\ref{fig:fusion}, we show the effect of combining both perspectives on UCF-101 and HMDB51. We fuse the action model of \cite{brattoli2020rethinking} with the object model of \cite{mettes2021object} and leave the final fusion with the model of \cite{pu2022alignment} for the state-of-the-art comparison.

\begin{table*}[t]
\centering
\begin{tabular}{l ccc @{\hskip 3\tabcolsep} ccc @{\hskip 3\tabcolsep} ccc}
\toprule
 & & & & \multicolumn{3}{c}{\textbf{UCF-101}} & \multicolumn{3}{c}{\textbf{HMDB51}}\\
 & K & A & O & 20 & 50 & 101 & 10 & 25 & 51\\
\midrule
\emph{Inductive} & & & & & & & & &\\
\midrule
\cite{hahn2019action2vec} & W & $\checkmark$ & & 36.5 {\scriptsize $\pm$ ---} & 22.1 {\scriptsize $\pm$ ---} & - & 40.1 {\scriptsize $\pm$ ---} & 23.5 {\scriptsize $\pm$ ---} & -\\
\cite{bishay2019tarn} & A,W & $\checkmark$ & & 42.7 {\scriptsize $\pm$ 5.4} & 23.2 {\scriptsize $\pm$ 2.9} & - & - & 19.5 {\scriptsize $\pm$ 4.2} & -\\
\cite{mishra2020zero} & A,W & $\checkmark$ & & - & 26.1 {\scriptsize $\pm$ 3.0} & - & - & 21.3 {\scriptsize $\pm$ 3.2} & -\\
%\cite{zhang2018visual} & & - & 28.8 {\scriptsize $\pm$ 5.7} & - & & - & 25.3 {\scriptsize $\pm$ 4.5} & -\\
%\cite{roitberg2018towards} & & - & - & - & & - & 25.7 {\scriptsize $\pm$ 3.5} & -\\
%Mettes and Snoek \cite{mettes2017spatial} & & 51.2 {\scriptsize $\pm$ 5.0} & 40.4 {\scriptsize $\pm$ 1.0} & 32.8 & & - & - & -\\
\cite{zhu2018towards} & W & $\checkmark$ & & 53.8 {\scriptsize $\pm$ ---} & 42.5 {\scriptsize $\pm$ ---} & 34.2 & - & - & -\\
\cite{mettes2021object} & W & & $\checkmark$ & 61.1 {\scriptsize $\pm$ ---} & 47.3 {\scriptsize $\pm$ ---} & 36.3 & - & - & -\\
\cite{bretti2021zero} & W & & $\checkmark$ & - & 45.4 {\scriptsize $\pm$ ---} & 39.3 & - & - & -\\
\cite{kim2021daszl} & A & $\checkmark$ & & - & 48.9 {\scriptsize $\pm$ 5.8} & - & - & - & -\\
\cite{brattoli2020rethinking} & W & $\checkmark$ & & - & 49.2 {\scriptsize $\pm$ ---} & 39.8 & - & 32.7 {\scriptsize $\pm$ ---} & 26.9\\
\cite{kerrigan2021reformulating} & W & $\checkmark$ & & - & 49.2 {\scriptsize $\pm$ ---} & 40.1 & - & 33.8 {\scriptsize $\pm$ ---} & 27.3\\
\cite{chen2021elaborative} & W & $\checkmark$ & $\checkmark$ & - & 51.8 {\scriptsize $\pm$ 2.9} & - & - & 35.3 {\scriptsize $\pm$ 4.6} & -\\
\cite{lin2022cross} & W & $\checkmark$ & & - & 54.7 {\scriptsize $\pm$ 2.3} & 41.2 & - & 39.3 {\scriptsize $\pm$ 3.5} & 30.6\\
\cite{pu2022alignment} & W & $\checkmark$ & & - & 58.0 {\scriptsize $\pm$ ---} & 46.8 & - & 39.0 {\scriptsize $\pm$ ---} & 31.7\\
\midrule
\emph{Transductive} & & & & & &\\
\midrule
\cite{xu2020transductive} & W & $\checkmark$ & & - & 30.0 {\scriptsize $\pm$ 1.8} & - & - & 29.8 {\scriptsize $\pm$ 2.2} & -\\
\cite{mandal2019out} & W & $\checkmark$ & & - & 38.3 {\scriptsize $\pm$ 3.0} & - & - & 30.2 {\scriptsize $\pm$ 2.7} & -\\
\cite{gao2019know} & W & & $\checkmark$ & - & 41.6 {\scriptsize $\pm$ 3.7} & - & - & 31.0 {\scriptsize $\pm$ 3.2} & -\\
\cite{zhuo2022zero} & W & & $\checkmark$ & - & \textbf{65.5} {\scriptsize $\pm$ 3.5} & - & - & 34.3 {\scriptsize $\pm$ 5.2} & -\\
%\cite{brattoli2020rethinking} $\star$ & & 00.0 {\scriptsize $\pm$ 0.0} & 00.0 {\scriptsize $\pm$ 0.0} & 39.2 & & 00.0 {\scriptsize $\pm$ 0.0} & 00.0 {\scriptsize $\pm$ 0.0} & 00.0\\
%\midrule
\rowcolor{Gray}
\textbf{This paper} & W & $\checkmark$ & & 68.9 {\scriptsize $\pm$ 6.3} & 57.3 {\scriptsize $\pm$ 4.0} & 49.4  & 54.8 {\scriptsize $\pm$ 8.7} & 41.4 {\scriptsize $\pm$ 4.3} & 33.4\\
%\textbf{This paper} (action) & 68.9 {\scriptsize $\pm$ 6.3} & 57.3 {\scriptsize $\pm$ 4.0} & 49.4  & 54.8 {\scriptsize $\pm$ 8.7} & 41.4 {\scriptsize $\pm$ 4.3} & 33.4\\
\rowcolor{Gray}
\textbf{This paper} & W & $\checkmark$ & $\checkmark$ & \textbf{71.4} {\scriptsize $\pm$ 6.4} & 62.1 {\scriptsize $\pm$ 4.4} & \textbf{51.4} & \textbf{58.3} {\scriptsize $\pm$ 5.4} & \textbf{42.8} {\scriptsize $\pm$ 4.1} & \textbf{33.9}\\
\bottomrule
\end{tabular}
\caption{\textbf{State-of-the-art comparison on UCF-101 and HMDB51} for different numbers of test actions. \reb{On all settings except UCF-101 with 50 classes, our approach obtains the highest zero-shot action classification scores and we expect further improvements when using Swin Transformers as employed by \citep{zhuo2022zero}.}}
\label{tab:sota1}
%$\circ$ denotes approaches that employ UCF-101 or HMDB51 seen actions for training and $\star$ denotes results from re-implementation. \psmm{observations and conclusions}}
\end{table*}

On UCF-101, we find that fusing both approaches has a clear, positive effect. Our results improve from 42.4\% (universal actions) and 31.6\% (universal objects) to 47.9\% when balancing both equally. When setting the fusion proportion to 0.3, the results can even be further improved to 48.9. Due to the zero-shot nature of our approach, we stick to an \emph{a priori} equal balance between both setups. Averaged over all fusion ratios, our approach provides a boost of 3.0 percent point compared to the baseline fusion. On HMDB51 with universal action models, adding our approach improves the results from 24.9\% to 28.1\%. The results are further improved to 29.4\% when fusing with universal object models. We conclude that both perspectives are complementary for zero-shot action recognition and their fusion benefits from our proposed transport.

\begin{table*}[t]
\centering
%\resizebox{0.9\textwidth}{!}{
\begin{tabular}{l @{\hskip 5\tabcolsep} ccccc @{\hskip 5\tabcolsep} ccccc}
\toprule
 & \multicolumn{5}{c}{\textbf{UCF Sports}} & \multicolumn{5}{c}{\textbf{J-HMDB}}\\
 & 0.1 & 0.2 & 0.3 & 0.4 & 0.5 & 0.1 & 0.2 & 0.3 & 0.4 & 0.5\\
\midrule
\emph{Inductive} & & & & & & & & & & \\
\midrule
\cite{jain2015objects2action} & 38.8 & 23.2 & 16.2 & 9.9 & 7.2 & - & - & - & - & -\\
\cite{mettes2017spatial} & 43.5 & 39.3 & 37.1 & 35.7 & 31.1 & 34.6 & 33.3 & 30.5 & 26.8 & 23.0\\
\cite{mettes2021object} & 47.3 & 43.0 & 40.7 & 37.9 & 33.1 & 37.3 & 37.1 & 33.9 & 31.0 & 26.7\\
\midrule
\emph{Transductive} & & & & & & & & & & \\
\midrule
\rowcolor{Gray}
\textbf{This paper} & \textbf{51.1} & \textbf{47.8} & \textbf{45.7} & \textbf{41.2} & \textbf{33.5} & \textbf{44.2} & \textbf{43.5} & \textbf{40.1} & \textbf{35.5} & \textbf{30.8}\\
\bottomrule
\end{tabular}
%}%
\caption{\textbf{State-of-the-art comparison on UCF Sports and J-HMDB} for localization with five overlap thresholds. Across datasets and thresholds, we obtain the highest scores, highlighting the effectiveness of our approach in the context of zero-shot spatio-temporal localization.}
\label{tab:sota2}
\end{table*}

\begin{figure*}
\centering
\includegraphics[width=0.975\textwidth]{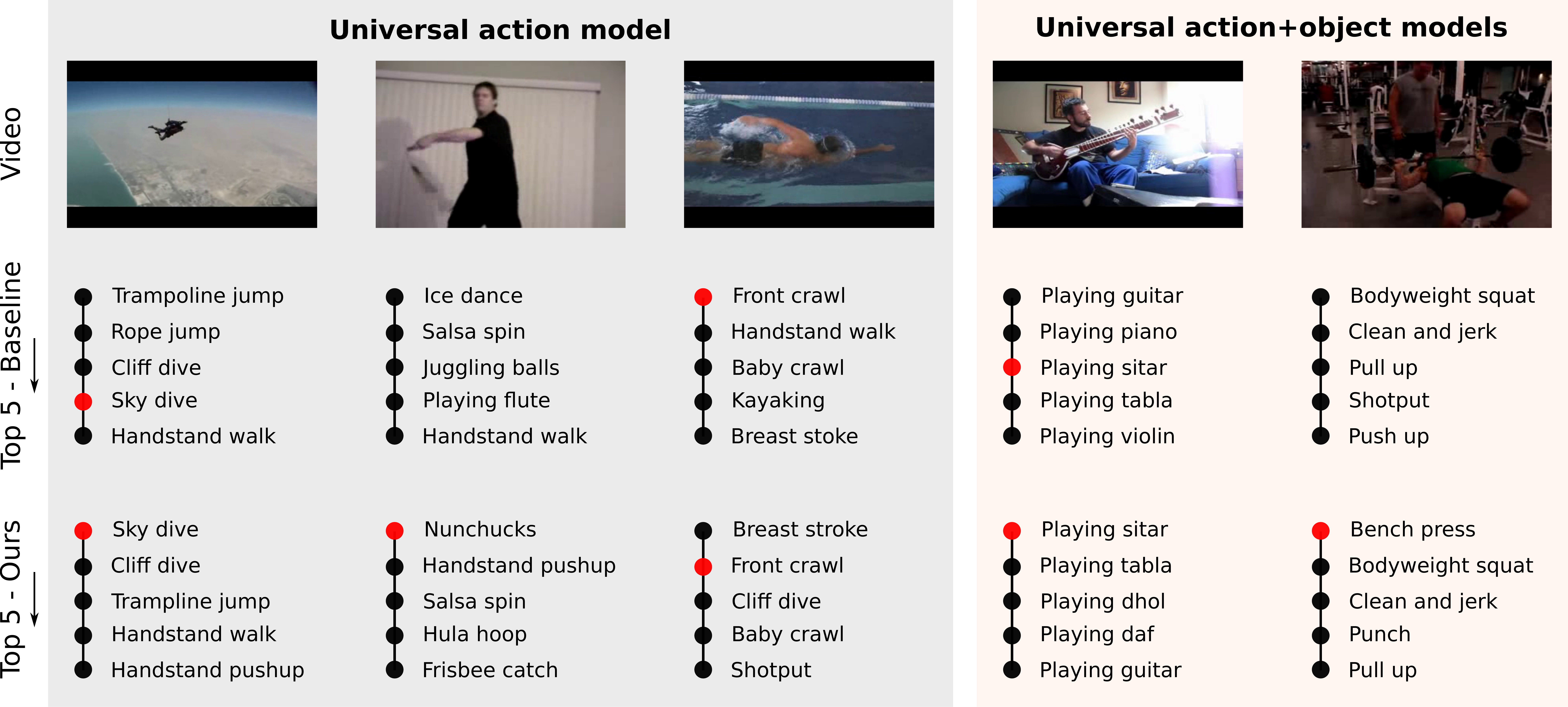}
\caption{\textbf{Qualitative examples} from UCF-101. \textbf{Columns 1-3:} Success and failure cases of our approach on the universal action model of \cite{brattoli2020rethinking}. Our approach can help to better align videos with unseen actions, as shown in the \emph{skydiving} and \emph{nunchuck} videos, but can fail for actions such as \emph{front crawl} due to confusion with a similar action like \emph{breast stroke}. \textbf{Columns 4-5:} Qualitative examples from the fusion of universal action and object models in our approach. In both cases, we are better able to classify fine-grained unseen actions by transferring dynamic knowledge about other actions and static knowledge about the objects used to perform these actions.}
\label{fig:qual}
\end{figure*}

%
%%%%% EXP 4
%
\subsection{Comparison to state-of-the-art}
\textbf{Zero-shot recognition.}
In Table~\ref{tab:sota1}, we compare our results on UCF-101 and HMDB51 to the state-of-the-art in zero-shot action recognition. Similar to other universal approaches, in the scenarios with random sub-selection of the test actions (20 and 50 for UCF-101, 10 and 25 for HMDB51) we do not use the remaining actions and their videos for network training. On both datasets, the current state-of-the-art is given by~\cite{lin2022cross} and \cite{pu2022alignment}. For the action number of our approach, we add our universal transport on top of the two action models used in Table~\ref{tab:addsota}. We note that when reproducing the results of \cite{pu2022alignment}, we obtained an average accuracy of 55.3\% on UCF-101 with 50 actions and 46.3\% with 101 actions. Our approach boosts these reproduced numbers to 57.3\% for 50 actions and 49.4\% for 101 actions. Across both datasets and dataset splits, we find that integrating universal prototype transport on both action and object models is important for zero-shot action recognition. On UCF-101 with 101 test actions, we obtain an accuracy of 51.4\%, the first result over the 50\% threshold in literature. On HMDB51 with 51 action we improve the results from 33.4\% to 33.9\%. We conclude that universal prototype transport is effective for zero-shot action recognition.

\reb{Table~\ref{tab:sota1} details comparisons with both complete testsets and smaller subsets. Averaged over the multiple runs, our approach is effective for both small and large dataset sizes. Compared to the baseline models used as starting point in our method, we find that the larger the testset, the higher the relative improvement, indicating that our approach benefits from richer semantics.}
\\\\
\textbf{Zero-shot localization.}
We also showcase the potential of our approach for zero-shot spatio-temporal action localization. Since our approach operates over entire videos, we start from the spatio-temporal tubes made publicly available by~\cite{mettes2021object}. For each tube, we simply add the score for each action from the entire video as given by our approach. In Table~\ref{tab:sota2}, we report the AUC scores for UCF Sports and J-HMDB. Across datasets and overlap thresholds, we find that the global scores from our approach boosts spatio-temporal localization. This is because the scores of our approach help to distinguish and rank tubes from different videos as they encode contextual information.

\reb{In conclusion, we find that for both zero-shot classification and spatio-temporal localization on all datasets, our approach provides consistent improvements, highlighting that universal prototype transport is effective across different collections of unseen actions.}

%
%%%%% EXP 5
%
\subsection{Qualitative analysis}
In Figure~\ref{fig:qual}, we show success and failure cases for our approach when applied to the universal action model of \cite{brattoli2020rethinking} and its fusion with the universal object model of \cite{mettes2021object}. Theses results reiterate the potential of combining action and object perspectives in zero-shot action recognition and the role of universal prototype transport in combining both views.

%% file: 6-conclusions.tex
\section{Conclusions}
In this work, we investigate a persistent limitation in current universal learning models for zero-shot action recognition, namely selection biases in the assignment of unseen actions to test videos. We introduce universal prototype transport to alleviate this limitation. Our approach consists of three stages: (i) finding an optimal transport mapping from unseen action prototypes to the projected test videos (in universal action models) or to object prototypes (in universal object models); (ii) obtaining a target prototype for each unseen action using the couplings from the hyperspherical optimal transport; and (iii) re-positioning the unseen actions along the geodesic spanned by the original and target prototypes. Empirical evaluation on four datasets shows the effectiveness of our approach for debiasing action assignments and for improving zero-shot recognition and localization as a result. Our approach is general and can be used to improve any universal model.
\\\\
\textbf{Data availability statement.} The pre-trained networks and meta-data of the universal action and object models are publicly available on the repos of \cite{brattoli2020rethinking}\footnote{github.com/bbrattoli/ZeroShotVideoClassification} and \cite{mettes2020shuffled}\footnote{github.com/psmmettes/shuffled-imagenet-bank}. All experiments are performed on publicly available datasets: UCF-101\footnote{www.crcv.ucf.edu/data/UCF101.php}, HMDB51\footnote{serre-lab.clps.brown.edu/resource/hmdb-a-large-human-motion-database/}, UCF Sports\footnote{www.crcv.ucf.edu/research/data-sets/ucf-sports-action/}, and J-HMDB\footnote{jhmdb.is.tue.mpg.de/}. The code for this paper will be made available on github.